\pdfoutput=1

\documentclass[11pt]{article}

\usepackage[preprint]{acl}

\usepackage{times}
\usepackage{latexsym}
\usepackage{amsmath}
\usepackage{amsfonts}
\usepackage{comment}

\usepackage[T1]{fontenc}
\usepackage{booktabs}

\usepackage[utf8]{inputenc}

\usepackage{microtype}

\usepackage{inconsolata}

\usepackage{graphicx}

\usepackage{listings}
\usepackage{xcolor}
\usepackage{multirow}

\title{RIRAG: Regulatory Information Retrieval and Answer
Generation}

\author{Tuba Gokhan\(^1\), Kexin Wang\(^2\), Iryna Gurevych\(^{1,2}\), Ted Briscoe\(^1\) \\
\(^1\)Mohamed Bin Zayed University of Artificial Intelligence (MBZUAI)\\
\(^2\)Ubiquitous Knowledge Processing Lab (UKP Lab)\\
Department of Computer Science and Hessian Center for AI (hessian.AI)\\
Technical University of Darmstadt }

\begin{document}
\maketitle
\begin{abstract}
Regulatory documents, issued by governmental regulatory bodies, establish rules, guidelines, and standards that organizations must adhere to for legal compliance. These documents, characterized by their length, complexity and frequent updates, are challenging to interpret, requiring significant allocation of time and expertise on the part of organizations to ensure ongoing compliance. Regulatory Natural Language Processing (RegNLP) is a multidisciplinary field aimed at simplifying access to and interpretation of regulatory rules and obligations.  We introduce a task of generating question-passages pairs, where questions are automatically created and paired with relevant regulatory passages, facilitating the development of regulatory question-answering systems. We create the ObliQA dataset, containing 27,869 questions derived from the collection of Abu Dhabi Global Markets (ADGM) financial regulation documents, design a baseline Regulatory Information Retrieval and Answer Generation (RIRAG) system and evaluate it with RePASs, a novel evaluation metric that tests whether generated answers accurately capture all relevant obligations while avoiding contradictions.

\end{abstract}

\section{Introduction}
Regulatory documents are formal semi-legal texts issued by governmental regulatory bodies, setting out rules, guidelines, and standards organizations must follow to ensure compliance. These documents typically cover a wide range of topics, from environmental and financial compliance to workplace safety and data protection. Characterized by their complex and precise language, they are designed to cover all potential scenarios, but can be challenging to interpret without specialized knowledge. Additionally, regulatory documents are frequently updated to reflect new laws, technological changes, or shifts in societal norms, necessitating continuous monitoring and adaptation by organizations to meet new compliance requirements. Compliance with these regulations requires considerable investment in terms of time and expertise. For example, between 1980 and 2020, the U.S. public and private sectors spent an estimated 292.1 billion hours complying with 36,702 regulations, accounting for approximately 3.2\% of total annual working hours \cite{10.1093/rfs/hhad001}. Moreover, compliance errors can lead to severe penalties, as, for example, demonstrated by the recent C\$7.5 million fine levied against the Royal Bank of Canada by FINTRAC for failing to report suspicious transactions \cite{rbc_fined_2023}.  Regulatory Natural Language Processing (RegNLP) is a recent multidisciplinary field designed to simplify access to and interpretation of regulatory rules and obligations, reducing errors and enhancing operational efficiency. 

Question-answer systems for regulatory documents enable stakeholders to better navigate complex rules and obligations. Stakeholders often need to extract precise obligations from these documents to address queries like, \textit{"What steps must we take when a transaction is suspected of involving money laundering?"}. However, due to the intricate structure of regulatory texts and the importance of adhering to semi-legal rules, generating accurate and comprehensive answers requires careful retrieval and synthesis of relevant obligations. These answers must include all necessary obligations while avoiding omissions or contradictions (see Appendix \ref{sec:sampleRIRAG}). This challenge highlights the need for advanced systems designed to automate and improve this process.

This paper makes several significant contributions to RegNLP:
\begin{enumerate}
    \item \textbf{Automated Question-Passages Generation}: We introduce a framework that employs Large Language Model (LLM) to automatically generate questions, with Natural Language Inference (NLI) integrated at the validation stage. This framework can be applied to specialized regulatory document collections to create datasets for the development and evaluation of question-answering and single passage or multi passages retrieval systems. 
    
    \item \textbf{ObliQA - Obligatory Question-Answer dataset}: We develop a multi-document, multi-passage QA dataset\footnote{\url{https://github.com/RegNLP/ObliQADataset}} specifically to facilitate research in RegNLP. It consists of 27,869 questions and their associated source passages, all derived from the full collection of regulatory documentation provided by Abu Dhabi Global Markets (ADGM)\footnote{\url{https://www.adgm.com/}}, the authority overseeing financial services in the Emirates of Abu Dhabi international financial centre and free zone.

    \item \textbf{RIRAG - Regulatory Information Retrieval and Answer Generation}: We introduce a task that first involves passage retrieval given a specific regulatory query, identifying all relevant obligations distributed across all regulatory documents. Following this, the task requires generating comprehensive answers based on the retrieved passages. These answers must synthesize information from multiple sources, ensuring that all pertinent obligations are included to fully address the regulatory query. The goal is to create responses that are both complete and concise, providing clear guidance without overwhelming the user with extraneous information.

    \item \textbf{RePASs - Regulatory Passage Answer Stability Score}: We propose RePASs\footnote{\url{https://github.com/RegNLP/RePASs}}, a novel reference-free metric designed to evaluate the quality of generated answers in regulatory compliance contexts. RePASs is tailored to the complex nature of regulatory texts, ensuring that all relevant obligations are accurately captured, fully reflected in the answers, and ideally devoid of contradictions. To effectively ensure that all obligations are covered, we developed an obligation classifier\footnote{\url{https://github.com/RegNLP/ObligationClassifier}} model fine-tuned on LegalBERT \cite{chalkidis-etal-2020-legal}, optimized for the detection of obligatory sentences within regulatory passages.
\end{enumerate}



\section{Related Work}
\label{sec:RelatedWork}

\subsection{Research on Regulatory NLP}

Recent RegNLP research has advanced significantly using various methodologies and datasets. \citet{10.1145/1165485.1165508} utilized information retrieval and extraction on accessibility regulations from the US and Europe, formatted into unified XML for semi-structured data handling. \citet{10.1007/978-3-540-87877-3_13} employed the Cerno framework for semantic annotation to automate extracting rights and obligations from U.S. HIPAA and Italian laws. \citet{chalkidis-etal-2018-obligation} introduced a hierarchical BiLSTM model for extracting obligations from legal contracts, demonstrating superior performance over traditional models. \citet{nair-etal-2018-towards} developed a pipeline for annotating global import-export regulations, enhancing compliance workflows. \citet{chalkidis-etal-2021-regulatory} proposed a two-step document retrieval process for EU/UK legislative compliance using BERT-based models, handling long document queries. \citet{9920030} presented an automated question-answering system using BERT, achieving high accuracy in identifying relevant texts from European regulations, including the General Data Protection Regulation.

Despite these advances, regulatory document retrieval faces challenges such as implicit obligations, multi-document queries, and cross-referenced legal clauses. Current systems often struggle to synthesize comprehensive answers when obligations are dispersed or fragmented, and they inadequately address contradictions and obligation coverage. Our work addresses these gaps by introducing a novel dataset and QA task tailored to the financial domain, built from the comprehensive collection of ADGM regulatory documents. Unlike previous datasets that focus on narrower aspects of compliance, our dataset captures a broader range of obligations across multiple documents. We also propose the RePASs evaluation metric, which goes beyond relevance to assess obligation coverage and contradictions, areas where existing metrics fall short.

\subsection{Retrieval-Augmented Generation}

RegNLP can benefit from advances in retrieval-augmented generation (RAG)~\cite{rag} and other related technologies to simplify access to regulatory documents and improve compliance. Regulatory documents present two key challenges for application systems: (1) frequent updates that can quickly render systems outdated, and (2) the complexity of specialized terminology and cross-document references that demand a comprehensive understanding of the entire document collection. RAG has significantly improved the accuracy, efficiency, and trustworthiness of LLMs by integrating external, contextually relevant and up-to-date information. Notable approaches include: Self-RAG~\cite{self-rag} enhances response quality by incorporating self-reflection mechanisms; RAFT~\cite{raft} optimizes domain-specific knowledge retrieval by filtering out irrelevant data and improving reasoning chains; RobustRAG~\cite{robustrag} and RQ-RAG~\cite{rq-rag} tackle challenges such as noisy retrieval results and ambiguous queries; PipeRAG~\cite{piperag} improves performance by reducing retrieval latency. However, these studies do not consider regulatory documents so we are interested in testing the ability of RAG methods for solving the QA task for regulatory questions.

\section{ObliQA: Obligation Based Question Answering Dataset}
\label{sec:ObliQA}
\begin{figure*}[!ht]
    \centering
    \includegraphics[width=1\linewidth]{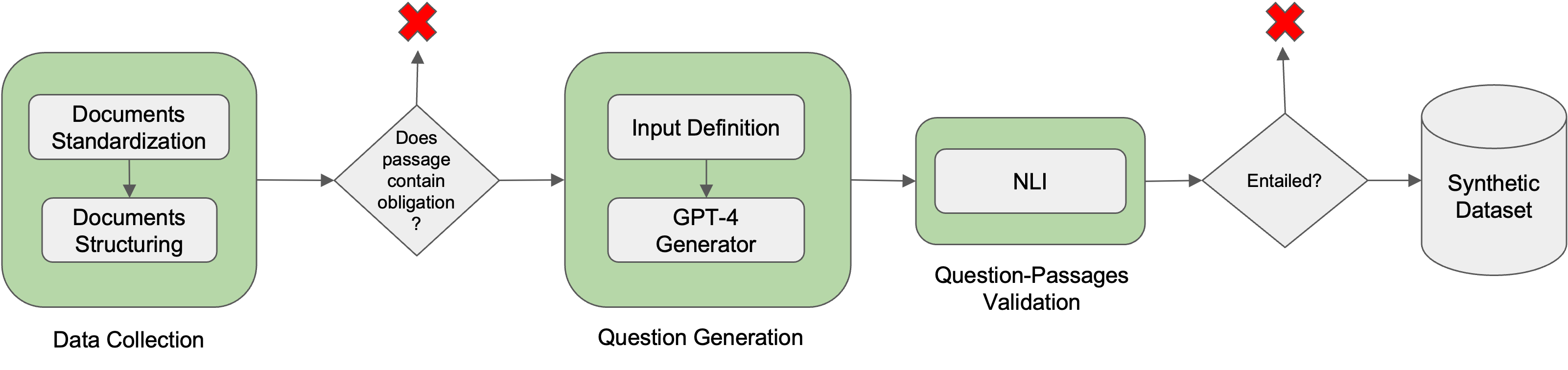}
    \caption{  AI-driven framework for generating question-passages pairs to create ObliQA: Obligation-Based Question Answering dataSet for regulatory compliance. This flowchart presents the process from data collection and standardization to question generation and NLI-based validation for constructing a synthetic dataset.}
    \label{fig:ObliQAFramework}
\end{figure*}

The ObliQA - The Obligation Based Question Answering Dataset creation process involves three main steps: Data Collection, Question Generation, and Question-Passages Validation via NLI, as illustrated in \autoref{fig:ObliQAFramework}. 

\begin{table*}[!ht]
\centering
\small
{%
\begin{tabular}{@{}l|c|cccccc@{}}
\toprule
               & \#Question     & 1 Passage      & 2 Passages     & 3 Passages    & 4 Passages   & 5 Passages   & 6 Passages   \\ \midrule
Group 1        & 7988          & 7988          &                &               &              &              &              \\
Group 2        & 7254          & 5432          & 1822           &               &              &              &              \\
Group 3        & 5365          & 3385           & 1434           & 546          &              &              &              \\
Group 4        & 4558           & 3057           & 1102           & 305           & 94          &              &              \\
Group 10       & 2704           & 1325           & 678           & 345           & 174          & 121          & 61          \\ \midrule
\textbf{Total} & \textbf{27869} & \textbf{21187} & \textbf{5036} & \textbf{1196} & \textbf{268} & \textbf{121} & \textbf{61} \\ \bottomrule
\end{tabular}%
}
\caption{Distribution of questions by number of passages in the ObliQA dataset. Groups indicate the number of input passages.}
\label{tab:question_distribution}
\end{table*}

\subsection{Data Collection}

The regulatory documents provided by ADGM consist of 40 documents totaling 640k words, each ranging from 30 to 100 pages. These documents are carefully structured with numbered clauses and subsections to maintain a hierarchical format. More details on document standardization and structuring processes can be found in the Appendix \ref{sec:DocumentStandardizationandStructuring}.


\subsubsection{Question Generation}
Regulatory documents contain extensive descriptions and titles, but for question generation, only passages containing obligations are considered essential, ensuring the questions are grounded in core compliance requirements.

\textbf{Input Definition:}

\textit{Inputs for Single Passage Questions:} Inputs for single passage questions isolate one specific source passage per question, simplifying the context and enhancing the relevance of the generated question.

\textit{Inputs for Multi-Passage Questions:} For broader queries, inputs encompass multiple source passages. Sixty-nine distinct topics related to finance and regulatory compliance (see Appendix \ref{sec:appendixKeywordTopicList} for a sample) guide the aggregation of related regulations into meaningful groups. These groups then form the basis for multi-passage questions by randomly sub-grouping a specified number of rules per question.

\textbf{GPT-4 Generator:} The \textit{gpt-4-turbo-1106} model is employed to generate both questions and answers.

For single passage questions, the prompt is:

\begin{quote}
\scriptsize
"Your task is to generate realistic and applied questions that pertain to the provided regulatory or compliance material. Ensure that the context implicitly contains the answer to the question."
\end{quote}

For multi-passage questions, the complexity increases, and the prompt is adjusted accordingly:

\begin{quote}
\scriptsize
"Imagine you are preparing a list of questions that a compliance officer at a company would ask regulatory authorities like the ADGM to clarify their understanding of compliance requirements and to ensure adherence to regulatory standards. Each question should be direct, practical, and structured to obtain detailed information on specific regulatory rules. Avoid formulations that imply the company is describing its own processes, and instead frame questions that seek clarity on regulatory expectations."
\end{quote}

\begin{table*}[!ht]
\centering
\small
{%
\begin{tabular}{@{}l|c|c|cccc@{}}
\toprule
               & \#Questions & \#Passages & Relevant & Irrelevant & Null  & Relevant but Indirect   \\ \midrule
Group 1        & 79          & 79         & 78          & 1               & 0              & 0              \\
Group 2        & 178         & 280        & 194          & 37              & 2              & 47              \\
Group 3        & 230         & 428        & 356           & 14              & 6              & 52              \\
Group 4        & 245         & 490        & 471           & 7               & 0              & 12          \\
Group 10       & 259         & 570        & 506           & 6               & 0              & 58          \\ \midrule
\textbf{Total} & \textbf{991} & \textbf{1847}  & \textbf{1605} & \textbf{65}    & \textbf{8}     & \textbf{169}   \\ \bottomrule
\end{tabular}%
}
\caption{Expert evaluation of passage relevance in the ObliQA dataset, categorized by groups.}
\label{tab:expert_evaluation}
\end{table*}
\subsection{Question-Passage(s) Validation via NLI}

Validating the generated question-passage(s) pairs using Natural Language Inference (NLI) involves assessing the semantic relationship between each question and its corresponding source passage(s) to ensure alignment \cite{10.1007/11736790_9}. The validation process uses the \textit{nli-deberta-v3-xsmall} model with passage(s) as the premise and question as the hypothesis. For each passage associated with a question, NLI is applied to evaluate the relationship. The validation results are categorized as follows.

\begin{itemize}
\item \textbf{Entailment}: The passage entail the questions being retained.
\item \textbf{Contradictions}: The passage that contradict the question is eliminated.
\item \textbf{Neutral}: The neutral passage is retained if the absolute difference between its neutral score and the entailment score is smaller than the difference between its neutral score and the contradiction score; otherwise, it is discarded.
\end{itemize}

The final dataset consists of approximately 21k questions linked to single passages and 5k questions linked to multiple passages. The distribution of questions among different groups, indicating the number of input passages, is detailed \autoref{tab:question_distribution}. The further distribution between the training, testing and development sets is shown in Appendix \ref{sec:distrubutionOBLIQA}. A sample of the dataset is provided in Appendix \ref{sec:appendixSampleObliQA}.

\subsection{Evaluation of ObliQA}
991 questions and their associated passages from the ObliQA dataset were evaluated by ADGM experts, as shown in Table \ref{tab:expert_evaluation}. All questions were labeled as "Expected Question," confirming alignment with real-world compliance scenarios. The evaluation covered 1,847 passages, of which 1,605 (86.91\%) were marked as relevant, 65 (3.52\%) as irrelevant, and 8 (0.43\%) as null. Additionally, 169 passages (9.14\%) were identified as relevant but did not directly answer the corresponding questions. These findings highlight the high quality and relevance of the ObliQA dataset for research in regulatory compliance.

\section{RePASs: Regulatory Passage Answer Stability Score}
\label{sec:RePASS}

We propose a reference-free evaluation metric, RePASs, to evaluate generated answers in regulatory compliance contexts. Inspired by recent work on answer attribution to specific sources \cite{bohnet2023attributed, yue-etal-2023-automatic}, this metric evaluates answers based on three key aspects:
\begin{itemize}
    \item Every answer sentence must be supported by at least one sentence in the source passage(s).
    \item The answer must not contain any sentences that contradict the information in the source passage(s).
    \item The answer must cover all the obligations present in the source passages, meaning that all regulatory obligations should be reflected in the answer.
\end{itemize}
This metric prefers answers that are not only accurate but also comprehensive in covering all relevant regulatory obligations. The metric includes the following steps:

\textbf{Step 1-} Generation of Entailment and Contradiction NLI Pair Matrices: We generate an NLI pair matrix by comparing each passage sentence (premise) with each answer sentence (hypothesis) using the \textit{cross-encoder/nli-deberta-v3-xsmall} \cite{he2021debertav3}. The model provides three outputs for each pair: probabilities for entailment, contradiction, and neutrality. These probabilities are organized into two matrices: the entailment matrix, which contains the entailment probabilities for each passage-answer sentence pair, and the contradiction matrix, which stores the contradiction probabilities for each sentence pair.

\textbf{Step 2-} Calculation of Entailment and Contradiction Scores: We reduce the size of the entailment and contradiction matrices by selecting the highest probability for each answer sentence across the passage sentences. Matrix reduction ensures that only the most relevant support from or conflict with the source passage is considered. The entailment score is computed by averaging the highest entailment probabilities across all answer sentences. Similarly, the contradiction score is obtained by averaging the highest contradiction probabilities. 
\begin{equation}
\displaystyle
E_s = \frac{1}{N} \sum_{i=1}^{N} \max_{j} P_{\text{entailment}}(p_j, a_i)
\end{equation}
\begin{equation}
\displaystyle
C_s = \frac{1}{N} \sum_{i=1}^{N} \max_{j} P_{\text{contradiction}}(p_j, a_i)
\end{equation}

Where \( E_s \) is the Entailment Score, and \( C_s \) is the Contradiction Score. \( N \) represents the total number of sentences in the generated answer. \( P_{\text{entailment}}(p_j, a_i) \) is the probability that the \( i \)-th sentence in the answer (\( a_i \)) is entailed by the \( j \)-th sentence in the source passage (\( p_j \)). The \(\max\) function selects the highest entailment probability for each answer sentence (\( a_i \)) across all sentences in the source passage (\( p_j \)). Similarly, \( \max_j \) selects the highest probability for each answer sentence across all sentences in the source passage. 

\textbf{Step 3-} Calculation of Obligation Coverage Score: The obligation coverage score evaluates how accurately the generated answer reflects the obligations present in the source passage(s). First, a dataset is created using an automated pipeline powered by the \textit{GPT-4-turbo-1106} model, which extracts and labels sentences from regulatory documents as either obligations or non-obligations using zero-shot learning. This labeled dataset is then used to fine-tune a LegalBERT-based classifier model \cite{chalkidis-etal-2020-legal}, which is optimized for detecting obligatory sentences in regulatory texts. Both the source passages and generated answers are tokenized into sentences to enable a detailed, sentence-level comparison for obligation coverage.


For each obligation sentence detected in the source passage, the system verifies its coverage in the generated answer. This is done by comparing the obligation sentences with those in the answer using the NLI model \textit{microsoft/deberta-large-mnli} \cite{he2021deberta}. If any sentence in the answer demonstrates an entailment score exceeding 0.7 when compared to an obligation sentence from the source passage(s), the obligation is considered covered. The obligation coverage score is calculated as the ratio of the number of covered obligations to the total number of obligations present in the source passage(s).

The formula for the Obligation Coverage Score is given by:
\begin{equation}
\displaystyle
OC_s = \frac{1}{M} \sum_{k=1}^{M} \mathbb{1} \left( \max_{l} P_{\text{entailment}}(o_k, a_l) > 0.7 \right)
\end{equation}

Where \( OC_s \) represents the Obligation Coverage Score, and \( M \) is the total number of obligation sentences in the source passage. \( P_{\text{entailment}}(o_k, a_l) \) denotes the probability that the \( k \)-th obligation sentence in the source passage (\( o_k \)) is entailed by the \( l \)-th sentence in the answer (\( a_l \)). The \(\max_{l}\) function selects the highest entailment probability for each obligation sentence across all sentences in the answer. The indicator function \( \mathbb{1} \) returns 1 if the highest entailment score exceeds 0.7, meaning the obligation is considered covered.

\textbf{Step 4-} Calculation of the Regulatory Passage Answer Stability Score: The final step integrates the three components into a single composite score. The score penalizes contradictions and normalizes the result between 0 and 1.
\begin{equation}
\displaystyle
RePASs = \frac{ \text{E}_{s} - \text{C}_{s} + \text{OC}_{s} + 1}{3}
\end{equation}

\subsection{Evaluation of RePASs}

\begin{table}[!ht]
\centering
\small
\begin{tabular}{llc}
\toprule
\textbf{Metric} & \textbf{Sub-metric} & \textbf{Score} \\
\midrule
\multicolumn{3}{l}{\textbf{Standard Evaluation Metrics}} \\
SummaC & - & 0.5890 \\
FactCC & - & 0.8571 \\
BERTScore & Precision & 0.6710 \\
 & Recall & 0.7449 \\
 & F1 & 0.7080 \\
BLEU & - & 0.8300 \\
ROUGE-1 & Precision & 0.8606 \\
 & Recall & 1.0000 \\
 & F1 & 0.9218 \\
ROUGE-2 & Precision & 0.8518 \\
 & Recall & 1.0000 \\
 & F1 & 0.9155 \\
ROUGE-L & Precision & 0.8606 \\
 & Recall & 1.0000 \\
 & F1 & 0.9218 \\
\midrule

\textbf{RePASs} & \(E_s\) & 0.7639 \\
 & \(C_s\) & 0.3336 \\
 & \(OC_s\) & 1.0000 \\
 & RePASs & 0.8101 \\
\bottomrule
\end{tabular}
\caption{Performance of evaluation metrics on the gold standard dataset}
\label{tab:RePASS_performance}
\end{table}

Traditional metrics like BLEU \cite{papineni2002bleu}, ROUGE \cite{lin-2004-rouge}, BERTScore \cite{zhang2020bertscoreevaluatingtextgeneration}, FactCC \cite{kryscinski-etal-2020-evaluating}, and SummaC \cite{laban-etal-2022-summac} are used to evaluate text generation but fall short in RIRAG contexts. BLEU and ROUGE focus on n-gram overlap and lexical similarity, neglecting factual accuracy and the completeness of regulatory obligations. BERTScore captures semantic similarity but lacks tools to assess whether answers cover all obligations or avoid contradictions. While FactCC ensures factual consistency, it doesn’t verify that all necessary regulatory obligations are included. SummaC evaluates consistency with the source text but doesn’t account for regulatory adherence or contradictions. In contrast, RePASs is tailored specifically to regulatory compliance. It evaluates answers based on entailment, contradiction, and obligation coverage, ensuring that responses not only maintain coherence but also comply with regulatory requirements.

To assess RePASs, we tested it on real-world questions using expert-provided obligations and answers from ADGM.  \autoref{tab:RePASS_performance} shows that while ROUGE-L achieves a high F1 score (0.9218), it doesn't ensure obligation coverage or the absence of contradictions. Similarly, BERTScore’s F1 score (0.7080) reflects semantic similarity but fails in regulatory compliance. FactCC’s score (0.8571) emphasizes factual consistency but doesn’t capture regulatory requirements. SummaC (0.5890) also falls short of the precision needed in these contexts. \autoref{tab:RePASS_performance} illustrates that RePASs aligns well with expert judgments, and the results provide an indicative upper bound for the performance of an automated system.

\section{Regulatory Information Retrieval and Answer Generation Task}
\label{sec:RIRAG}

The Regulatory Information Retrieval and Answer Generation (RIRAG) task includes two key subtasks: passage retrieval and answer generation. In the passage retrieval subtask, the system retrieves all relevant obligations from regulatory documents in response to a given question. The goal is to ensure that all pertinent obligations are identified and collected. In the answer generation subtask, the system synthesizes the retrieved information into a precise, informative response that comprehensively addresses the question. In this section, we describe a baseline approach for RIRAG, setting a standard that will support future research in the field.


\subsection{Passage Retrieval}

We experiment with six retrieval models, which are (1) \textbf{BM25}~\cite{bm25}: the traditional lexical-based model;
(2) \textbf{DRAGON+}~\cite{dragon_plus}: the State-of-the-Art (SotA) single-vector dense retriever model fine-tuned on MS MARCO~\cite{msmarco}; (3) \textbf{SPLADEv2}~\cite{spladev2}: the SotA neural sparse retriever fine-tuned on MS MARCO; (4) \textbf{ColBERTv2}: the SotA multi-vector dense retriever model finetuned on MS MARCO; (5) \textbf{NV-Embed-v2}~\cite{nv_embed}: the SotA single-vector dense retriever model fine-tuned on multiple text-embedding datasets, achieving the best\footnote{As of Aug 30, 2024.} performance on Massive Text Embedding Benchmark (MTEB)~\cite{muennighoff-etal-2023-mteb}; (6) \textbf{BGE-EN-ICL}~\cite{bge_embedding}: the SotA single-vector dense retrieval model fine-tuned on multiple text-embedding datasets, achieving the second-best\footnote{As of Aug 30, 2024.} performance on MTEB. 

DRAGON+, SPLADEv2 and ColBERTv2 represent the three main architectures for neural retrieval, achieving SotA effectiveness with limited training resources and parameters (all around 110M parameters). NV-Embed-v2 and BGE-EN-ICL are the best-performing dense retrievers when the training resources are extensive and the parameters are largely scaled up (7.1B and 7.9B parameters, respectively). Additionally, BGE-EN-ICL supports In-Context Learning (ICL)~\cite{icl} for text embeddings, providing an option for obtaining task-adapted model with a low cost via adding example demonstrations in the model input.

Since the retrieval task in RIRAG searches forretrieve passages from a corpus of long documents, we also consider it as a Document-Aware Passage Retrieval (DAPR)~\cite{wang-etal-2024-dapr} task to jointly model the document context and passage retrieval. To this end, we apply \textbf{rank fusion} to linearly fuse the passage ranking by the neural or BM25 retrievers and the document ranking by the BM25 retriever. \citet{wang-etal-2024-dapr} shows that this approach yields a consistent improvement on retrieval tasks across different domains. The details are available in Appendix \ref{sec:rank_fusion}. The counterpart without rank fusion is denoted as \textbf{passage-only}.

For evaluating the retrieval module in RIRAG, we use recall@10 as the main evaluation metric, as we rely on the retrieval module to cover the relevant information as much as possible while leaving the burden of noise filtering to the answer-generation module. This approach is in line with LongRAG~\cite{jiang2024longragenhancingretrievalaugmentedgeneration}. To additionally assess the ranking ability within the top-10 passages, MAP@10 (Mean Averaged Precision) is also used for diagnostic purposes.

For the experimental setup, we truncate the queries and the passages to 512 tokens. In rank fusion, we fuse the ranking of top-100 passages and the ranking of all the documents. The fusion weight on the document side is set to 0.1. The ranking scores are normalized (cf.~\autoref{eq:normalization}) with respect to the top-100 passages into the range of $[0, 1]$ for filtering the passages conveniently in the answer-generation step. For BGE-EN-ICL, we consider the scenarios of zero-shot, 1-shot, 3-shot and 5-shot. The ICL examples are obtained via random sampling from the training set of ObliQA. According to their recommended usage, the instructions/query prefixes\footnote{These models do not support instructions for the passage input.} in NV-Embed-v2 and BGE-EN-ICLL are ``Given a question, retrieve passages that answer the question'' and ``Given a financial-regulation query, retrieve relevant passages that answer the query'', respectively.

\subsection{Answer Generation}
The answer generation process begins once 10 relevant passages have been retrieved. In our post-retrieval stage, we implement a score-based filtering approach with a threshold of 0.2 to identify significant drops in relevance between consecutive passages. We also enforce a minimum score criterion of 0.7, ensuring that only the most relevant passages are considered for generating answers. Based on these filtered passages, we use the \textit{gpt-4-turbo-1106} model to generate comprehensive answers. The model follows a tailored prompt designed to simulate the role of a regulatory compliance assistant, integrating all relevant obligations and best practices from the passages into a cohesive response. The prompt reads:
\begin{quote}
    \scriptsize
    "You are a regulatory compliance assistant. Provide a detailed answer for the question that fully integrates all the obligations and best practices from the given passages. Ensure your response is cohesive and directly addresses the question. Synthesize the information from all passages into a single, unified answer."
\end{quote}
For evaluation, the responses are organized in JSON format, as shown in the Appendix \ref{sec:SampleRePASsData}.

\subsection{Results}

\subsubsection{Passage Retrieval Results}
The passage retrieval results on the test split of ObliQA are shown in~\autoref{tbl:retrieval_obliqa_test}. We find the models of DRAGON+, SPLADEv2 and ColBERTv2, which are fine-tuned with limited training data (i.e. MS MARCO only) and limited parameters, generalize poorly, achieving worse or on-par recall@10 and MAP@10 than simple BM25. ColBERTv2, the best model in this group, outperforms BM25 by 0.6 points MAP@10 and 1.1 points recall@10. On the other hand, the models of NV-Embed-v2 and BGE-EN-ICL, which are fine-tuned with more training data and more parameters, outperform BM25 significantly in all the settings. Compared with BM25, recall@10 and MAP@10 are improved by up to 3.4 points and up to 2.1 points, respectively. These results show that scaling the retrievers with more training data and parameters is necessary for good generalization on the retrieval task in RIRAG. We find ICL is important for adapting the neural models to this new task, while the advantage of including more examples in the demonstration is very marginal. 
Lastly, we find that rank fusion, which considers the document context during passage retrieval, improves the performance only marginally. We assume this is because the query generation process only considers the target passages as input, resulting in fairly self-contained query-evidence pairs.

\begin{table}[!ht]
\small
\centering
\resizebox{\linewidth}{!}{
\begin{tabular}{lcccc} 
\toprule
\multirow{2}{*}{\textbf{Model}} & \multicolumn{2}{c}{\textbf{Passage-only}} & \multicolumn{2}{c}{\textbf{Rank fusion}}  \\ 
                                & \textbf{R@10} & \textbf{M@10}      & \textbf{R@10} & \textbf{M@10}      \\ 
\midrule
BM25&76.1&62.4&76.4&62.5\\
\hline
DRAGON+&74.3&58.1&74.5&58.2\\
SPLADEv2&75.7&60.4&75.9&60.7\\
ColBERTv2&77.7&63.0&77.7&63.2\\
\hline
NV-embed-v2&78.3&63.5&78.2&63.6\\
BGE-EN-ICL (0-shot)&77.1&62.4&77.4&62.5\\
BGE-EN-ICL (1-shot)&79.2&64.1&79.3&64.2\\
BGE-EN-ICL (3-shot)&\textbf{79.5}&64.1&\textbf{79.5}&64.2\\
BGE-EN-ICL (5-shot)&79.3&\textbf{64.4}&\textbf{79.5}&\textbf{64.5}\\
\bottomrule
\end{tabular}}
\caption{Results of the retrieval task on the test dataset. R@10 and M@10 represent Recall@10 and MAP@10, respectively.}
\label{tbl:retrieval_obliqa_test}
\end{table}

\subsubsection{Answer Generation Results}
\autoref{tab:answer_generation_results-test-set} illustrates the performance differences between BM25(passage-only)+GPT-4 and BM25(rank fusion)+GPT-4 used in our answer generation task. Similar to the retrieval task results, where rank fusion showed slight performance improvements, we observe that BM25 (rank fusion) + GPT-4 outperforms BM25 (passage-only) + GPT-4 across all metrics. Specifically, rank fusion achieves higher E\textsubscript{s} (0.320 vs. 0.308), C\textsubscript{s} (0.131 vs. 0.123), OC\textsubscript{s}(0.222 vs. 0.214), and RePASs (0.470 vs. 0.466). This highlights the connection between better passage retrieval and improved answer generation in regulatory tasks.

\begin{table}[!ht]
    \centering
    \small
    \resizebox{\columnwidth}{!}{%
    \begin{tabular}{lcccc}
       \toprule
       \textbf{Method}        & \textbf{E\textsubscript{s}} & \textbf{C\textsubscript{s}} & \textbf{OC\textsubscript{s}} & \textbf{RePASs} \\
       \midrule
       BM25(passage-only)+GPT-4          & 0.308            & 0.123               & 0.214                     & 0.466       \\
       BM25(rank fusion)+GPT-4   & \textbf{0.320}            & \textbf{0.131}               & \textbf{0.222}                     & \textbf{0.470}       \\
       \bottomrule
    \end{tabular}%
    }
    \caption{Results of the answer generation task using RePASs on the ObliQA test dataset. E\textsubscript{s}, C\textsubscript{s}, OC\textsubscript{s} and RePASs represent Entailment, Contradiction, Obligation Coverage and RePAS score, respectively. }

    \label{tab:answer_generation_results-test-set}
\end{table}

\section{Conclusion}
\label{sec:Conclusion}
In conclusion, this paper advances the field of RegNLP by introducing research infrastructure and metrics aimed at improving the overall precision of regulatory compliance. The automated question-passage generation framework and the ObliQA dataset contribute to the study of the retrieval and understanding of regulatory information. Furthermore, the regulatory information retrieval and answer generation task, supported by the novel RePASs evaluation metric, provides a framework for improving on our baseline approach.

In the future, the field of RegNLP could expand with the integration of new tasks. For instance, \textit{Summarization and Simplification} methods could be used to develop tools to make complex regulatory texts simpler and more accessible to nonexperts. \textit{Automated Compliance Checking} could leverage cross-document analysis techniques to improve regulatory adherence by comparing an organization's internal compliance documents with regulatory documents. \textit{Regulatory Gap Analysis} could focus on identifying vagueness, ambiguity or contradictions in regulations to aid in their refinement and ongoing compliance when regulations are added or updated. The continued development of these tasks will equip RegNLP to better serve the needs of regulatory bodies and regulated organizations alike, making compliance a more streamlined, reliable and efficient process.

\section{Limitations}
\label{sec:Limitations}

One significant limitation we encountered is the scarcity of publicly available datasets. Regulatory compliance often involves handling highly sensitive and confidential information, such as personal data or financial records, which presents challenges in terms of data privacy and security. To address this, we created a semi-synthetic dataset. During the validation phase, we faced difficulties stemming from multidisciplinary complexity and shortage of experts. However, experts evaluated a small subset of the dataset and confirmed that 100\% of the generated questions met expectations, while 86\% of the answer passages were deemed relevant to the questions.

Another key limitation is the lack of standardized formats for regulatory documents. Regulatory documents vary significantly across industries, regions, and even within the same organization, making it difficult to develop models capable of generalization.In this work, standardizing 40 documents required over 300 person-hours of effort. Additionally, the handling of tables and figures presented further challenges; tables were semi-automatically converted to text format, and figures were interpreted using the GPT-4 model.

Furthermore, the absence of comprehensive benchmarks is another limitation. There is no standardized benchmarking system available to evaluate the performance of RegNLP models across various regulatory domains, making it challenging to assess the effectiveness of different approaches or track progress in the field. Although we proposed the RePASs metric, we were unable to compare it with other models due to the lack of obligation coverage in existing benchmarks. Nevertheless, on genuine gold standard data from ADGM, RePASs demonstrated its suitability for RIRAG tasks.

\section*{Acknowledgement}
We would like to express our gratitude to the Financial Services Regulatory Authority (FSRA)'s Applied Innovation \& Research Team within ADGM for providing data and annotations. We also gratefully acknowledge ADGM's support through a grant, enabling access to OpenAI GPT models via the Azure cloud, which contributed significantly to the technical development of this research.

\bibliography{custom}

\onecolumn
\appendix
\section{Appendix}
\label{sec:appendix}

\subsection{Illustrative Example of RIRAG for Compliance Queries}
\label{sec:sampleRIRAG}

\begin{figure*}[ht]
    \centering
    \includegraphics[width=1\linewidth]{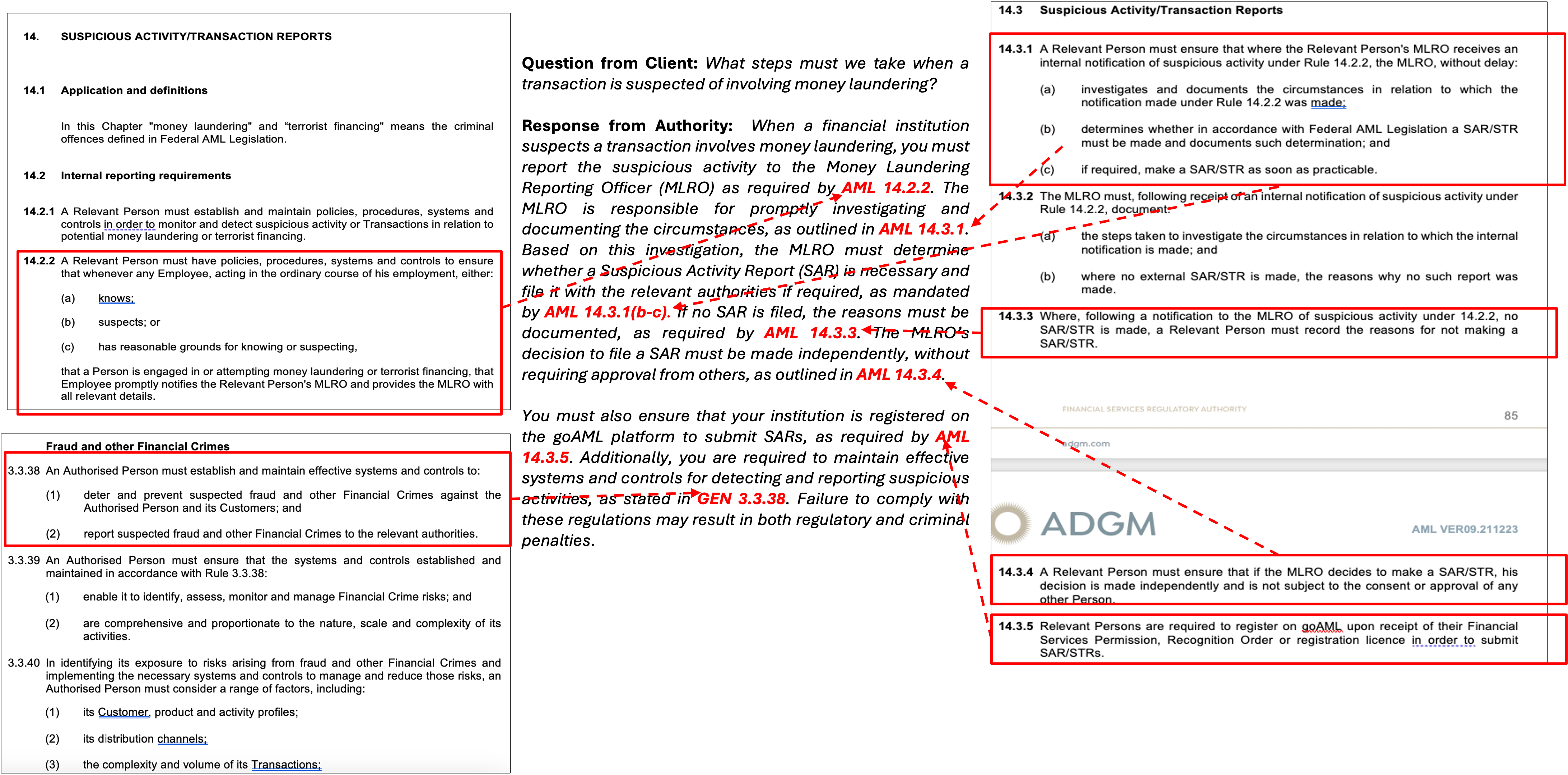}
    \caption{An example of a comprehensive response by retrieving and synthesizing relevant passages from regulatory documents in response to a client question}
    \label{fig:sampleRIRAG}
\end{figure*}

\autoref{fig:sampleRIRAG} illustrates a RIRAG task. In response to a client’s query about handling suspicious financial transactions, the system retrieves key regulations from several passages across the Anti-Money Laundering (AML) and General (GEN) sections of regulatory documents. By synthesizing these passages, the system generates a concise, comprehensive answer, streamlining the institution's compliance with complex regulations.

\subsection{Document Standardization and Structuring}
\label{sec:DocumentStandardizationandStructuring}

\textbf{Documents Standardization:}
To ensure the integrity of the dataset, the documents underwent a meticulous semi-automatic restructuring process. This step is critical due to the diverse formats employed by different departments in the preparation of regulations. Each document is parsed and transformed into .txt format. Tables within the documents are delineated with \textit{\textbackslash Table Start} and \textit{\textbackslash Table End} tags, while graphical content is transcribed into text using manually-corrected GPT-4 output and encapsulated with \textit{\textbackslash Figure Start} and \textit{\textbackslash Figure End} tags.

\textbf{Structuring the Data:}
Subsequently, the standardized .txt files are converted into a structured JSON format for data manipulation and analysis. Below is an example of the JSON structure:

\begin{lstlisting}
{
    "ID": "2230472f-a9d4-4b81-842f-964c0653f8e7",
    "DocumentID": 20,
    "PassageID": "1.1",
    "Passage": "This Guidance has been produced to help legal practitioners in ADGM to understand the intention behind the Application of English Law Regulations 2015 (the 'Application Regulations') and how English law has been implemented in ADGM..."
}
\end{lstlisting}

The \texttt{PassageID} is used to represent the hierarchical information of the passages within the documents.

\subsection{Keyword-Topic List for ObliQA Dataset}
\label{sec:appendixKeywordTopicList}
\begin{lstlisting}
topics_keywords = {
    "AML Compliance": [
        "money laundering", "compliance program", "regulatory requirements"
    ], 
    "Anti-Money Laundering": [
        "AML","money laundering", "KYC", "know your customer", "financial crime", "terrorist financing",
        "due diligence", "suspicious activity", "compliance program"
    ],
    "Audit and Monitoring": [
        "audit", "compliance monitoring", "internal audits", "external audits"
    ],
    "Blockchain Technology": [
        "blockchain", "blockchain technology", "smart contract", "tokenization"
    ],
    "Blockchain-based Securities": [
        "blockchain-based securities", "blockchain technology in securities"
    ],
    ....
    "Virtual Asset Regulation": [
        "virtual assets", "crypto assets", "digital asset regulation", "virtual asset service providers", "VASP", "crypto exchanges", "crypto custodians", "ICO regulations", "token classifications"
    ]
}

\end{lstlisting}

\subsection{Distribution of questions in the ObliQA dataset}
\label{sec:distrubutionOBLIQA}

\begin{table*}[!ht]
    \centering
    \small
    \begin{tabular}{@{}l|c|cccccc@{}}
    \toprule
         & \#Question     & 1 Passage      & 2 Passages     & 3 Passages    & 4 Passages   & 5 Passages   & 6 Passages   \\ \midrule
        train & \textbf{22295} & 16946 & 4016  & 975 & 202 & 100 & 56\\
        test & \textbf{2786} & 2126 & 506 & 105 & 36 & 9 & 4\\
        development & \textbf{2888} & 2215 & 514 & 116 & 30 & 12 & 1\\ \bottomrule
    \end{tabular}
    \caption{Distribution of questions in the ObliQA dataset across training, testing, and development sets , categorized by the number of associated passages.}
    \label{tab:question_distribution2}
\end{table*}

\subsection{JSON Data Examples from the ObliQA Dataset}
\label{sec:appendixSampleObliQA}
\begin{lstlisting}
[
  {
    "QuestionID": "739921c1-385a-4735-a052-dee9fba73602",
    "Question": "What are the key compliance indicators that a Fund Manager should monitor to confirm that a Passported Fund is being managed and operated within its constitutional framework and applicable ADGM legislation?",
    "Passages": [
      {
        "DocumentID": 16,
        "PassageID": "Part 3.6.(2)",
        "Passage": "Each Reporting UAE Financial Institution shall establish and implement appropriate systems and internal procedures to enable its compliance with the Cabinet Resolution and these Regulations."
      },
      {
        "DocumentID": 5,
        "PassageID": "6.1.2",
        "Passage": "The Fund Manager of a Passported Fund must:\n(a)\tensure that the Passported Fund is at all times managed and operated in compliance with its constitution, in accordance with applicable ADGM legislation, and with these Rules; and\n(b)\tmaintain, or cause to be maintained, a Unitholder register for the Passported Fund."
      }
    ]
  },
  {
    "QuestionID": "8dc451a4-ec55-4fa5-abce-b0b8764a9338",
    "Question": "Can the Regulator provide case studies or examples where the conduct of an Approved Person has been deemed compliant or non-compliant with the Principles, to help clarify the practical application of these rules?",
    "Passages": [
      {
        "DocumentID": 6,
        "PassageID": "PART 5.13.3.9.Guidance.3",
        "Passage": "The onus will be on the Regulator to show that he is culpable, taking into account the standard of conduct required under the principle in question. In determining whether or not the particular conduct of an individual complies with the principles, the Regulator will take into account whether that conduct is consistent with the requirements and standards relevant to an individual's role and the information available to him."
      },
      {
        "DocumentID": 7,
        "PassageID": "2.3.1.Guidance.3.",
        "Passage": "In those circumstances, the onus is on the Regulator to show that the Approved Person is culpable, taking into account the standard of conduct required under the Principle in question. In determining whether or not the particular conduct of an Approved Person complies with the Principles for Approved Persons, the Regulator will take account of whether that conduct is consistent with the requirements and standards relevant to their Authorised Person, their own role and the information available to them.\n"
      }
    ]
  }
]
\end{lstlisting}

\subsection{JSON Data Example for Evaluation of Answer Generation Step}
\label{sec:SampleRePASsData}
\begin{lstlisting}
{
    "QuestionID": "12345678-abcd-1234-efgh-ijklmnopqrst",
    "Question": "What are the regulatory requirements for data encryption in financial services?",
    "RetrievedPassages": [
        "Financial institutions must encrypt sensitive data at rest and in transit. Key requirements include the use of AES-256 encryption for data at rest and TLS 1.2 or higher for data in transit."
    ],
    "Answer": "Regulatory requirements for data encryption in financial services mandate using AES-256 encryption for data at rest and TLS 1.2 or higher for data in transit to ensure security of sensitive information.",
    "RetrievedIDs": [
        "9abc1234-def5-6789-ghij-klmnopqrstuv"
    ]
}
}
\end{lstlisting}

\subsection{Rank Fusion}
\label{sec:rank_fusion}
Rank fusion fuses the relevance scores from a BM25 retriever and a neural retriever. We compute the fusion as the convex combination of the normalized relevance scores~\cite{dense-retrievers-require-interpolation}:
\begin{align*}
    &s_{\mathrm{convex}}(q, p, d)=\alpha \hat{s}_{\mathrm{BM25}}(q, p) + (1- \alpha) \hat{s}_{\mathrm{neural}}(q, d),
\end{align*}
where $\alpha\in[0,1]$ is the fusion weight, and $\hat{s}_{\mathrm{BM25}}$/$\hat{s}_{\mathrm{neural}}$ represents the normalized BM25/neural-retrieval relevance score, respectively. The normalization for a relevance score $s$ is calculated as:
\begin{equation}
    \label{eq:normalization}
    \hat{s}(q, c) = \frac{s(q, c) - m_q}{ M_q - m_q},    
\end{equation}
where $c$ represents the passage/document candidate, $m_q$ and $M_q$ are the min. and max. relevance scores of the top candidates for $q$, respectively. For any misaligned candidates, a zero score is taken for the candidate-missing side. In this work, the fusion is applied between the BM25 document rankings and the neural passage rankings of a certain cut-off.

\end{document}